\documentclass[journal]{IEEEtran}
\usepackage{amsmath,amsfonts}
\usepackage{algorithmic}
\usepackage{algorithm}
\usepackage{array}
\usepackage{subcaption}
\usepackage{textcomp}
\usepackage{stfloats}
\usepackage{hyperref}
\usepackage{url}
\usepackage{verbatim}
\usepackage{subcaption}
\usepackage{graphicx}
\usepackage{soul}
\usepackage{makecell}
\usepackage{cite}
\usepackage{booktabs}
\usepackage{multirow}
\usepackage{threeparttable}
\usepackage{adjustbox}
\usepackage[listings,breakable,skins]{tcolorbox}
\usepackage{ulem}
\usepackage{wrapfig}
\usepackage{color}
\usepackage{pdfpages}

\newcommand{\x}{\boldsymbol{x}}
\newcommand{\X}{\mathcal{X}}

\newtcblisting{prompttemplate}[1]{
  title={#1},
  breakable,
  boxrule=0.5mm,
  listing only,
  listing options={
    basicstyle=\ttfamily,
    breaklines=true,
    columns=fullflexible,
    keepspaces=true,
  }
}
\newtcblisting{prompttemplateWhite}[1]{
  title={#1},
  breakable,
  boxrule=0.5mm,
  colframe=gray!50!black,
  colback=white,
  coltitle=white, 
  fonttitle=\scshape,
  listing only,
  listing options={
    basicstyle=\ttfamily\footnotesize,
    breaklines=true,
    columns=fullflexible,
    keepspaces=true,
  }
}

\begin{document}

\title{MiniOpt: Reasoning to Model and Solve General Optimization Problems with Limited Resources}
\author{Ke Zhao, Zixiang Di, Hong Qian, Yaolin Wen, Xiang Shu, Qitao Shi, Bingdong Li, Xingyu Lu,\\ Xiangfeng Wang, Jun Zhou, Ke Tang~\IEEEmembership{Fellow,~IEEE} and Yang Yu~\IEEEmembership{Member,~IEEE}
\thanks{This work was supported in part by National Natural Science Foundation of China (Grant No. 62476091). (Corresponding author: Hong Qian.)}
\thanks{Ke Zhao, Zixiang Di, Yaolin Wen, Bingdong Li, and Xiangfeng Wang are with East China Normal University, Shanghai 200062, China (E-mail: kezhao@stu.ecnu.edu.cn).}
\thanks{Hong Qian is with East China Normal University and Shanghai Innovation Institute, Shanghai 200062, China. (E-mail: hqian@cs.ecnu.edu.cn).}
\thanks{Xiang Shu, Qitao Shi, Xingyu Lu, and Jun Zhou are with AntGroup, Hangzhou 310000, China. (E-mail: sing.lxy@antgroup.com).}
\thanks{Ke Tang is with Southern University of Science and Technology, Shenzhen 518055, China. (E-mail: tangk3@sustech.edu.cn).}
\thanks{Yang Yu is with Nanjing University, Nanjing 210023, China. (E-mail: yuy@nju.edu.cn).}

\thanks{This work has been submitted to the IEEE for possible publication. Copyright may be transferred without notice, after which this version may no longer be accessible. \textcopyright\ 2026 IEEE.}}

\markboth{Journal of \LaTeX\ Class Files}%
{Shell \MakeLowercase{\textit{et al.}}: A Sample Article Using IEEEtran.cls for IEEE Journals}

\maketitle

\begin{abstract}

Achieving strong optimization generalization across diverse optimization problems while requiring limited training resources remains a challenging problem for optimization-oriented large language models (LLMs). Existing learning-based methods, however, typically require large-scale supervised data, expensive intermediate step verification, and extensive computational resources, while exhibiting limited optimization generalization across diverse optimization structures. In this paper, we propose \textsc{MiniOpt}, a lightweight optimization-specialized LLM trained through a reasoning-to-model-and-solve paradigm and a two-stage training framework with reinforcement learning with verifiable rewards (RLVR). MiniOpt encourages structure-oriented optimization learning by reinforcing the completeness of optimization formulations and optimization outcomes. Extensive experiments on eight optimization benchmarks demonstrate that MiniOpt consistently outperforms existing learning-based optimization models while requiring substantially fewer parameters and training data. To further understand optimization generalization, we introduce a complementary parameter-decoupled evaluation protocol that disentangles optimization structure understanding from numerical parameter processing. The results reveal that MiniOpt preserves optimization capability under parameter-decoupled settings significantly better than previous optimization-specialized models, suggesting that its performance gains primarily originate from stronger optimization structure understanding. These findings demonstrate an effective pathway toward efficient and generalizable optimization-specialized LLMs.
The code is available at \href{https://github.com/Hsiang-1/MiniOpt}{https://github.com/Hsiang-1/MiniOpt}.

\label{abstract}
\end{abstract}


\section{Introduction}
\IEEEPARstart{O}{ptimization} problems are ubiquitous in real-world scenarios, profoundly affecting diverse domains, such as industrial production and transportation planning\cite{9826438,li2025arsautomaticroutingsolver}. While traditional optimization solvers are efficient, their application heavily relies on expert knowledge, requiring the manual conversion of problems described in natural language into precise mathematical formulations and executable code. This process is both time-consuming and non-generalizable. The rise of LLMs has opened new pathways for the automated modeling and solving of optimization problems using natural language descriptions\cite{deng2024cafa,sun2024autosatautomaticallyoptimizesat, li2025autopbollmpoweredoptimizationlocal}, making them more applicable to application scenarios. Representative work like LLMOPT\cite{JiangShu2025llmopt}, BPP-Search\cite{wang-etal-2025-bpp}, and Text2Zinc\cite{singirikonda2025text2zinccrossdomaindatasetmodeling} significantly advances this field by parsing natural language descriptions into structured formulation that provide a unified representation of optimization problems, and subsequently generating solver code efficiently.


However, deploying such LLM-based approaches faces three critical bottlenecks. First, ensuring accurate text generation with a small-scale model requires Supervised Fine-Tuning (SFT) with a large amount of high-quality training data\cite{wu-etal-2025-training, lu2025optmathscalablebidirectionaldata}. However, obtaining such data often requires considerable time and effort, and is prone to errors\cite{xiao2025surveyoptimizationmodelingmeets}. 
Second, it is challenging to verify whether the result generated meets the requirements\cite{zadorojniy2025agentbasedframeworkautomaticvalidation,zhai2025equivamap}. The non-verifiability of this task also exposes the limitations of the learning-based solving paradigm. Previous methods often incorporate reflection or debugging mechanisms\cite{pmlr-v235-ahmaditeshnizi24a,li2025optbenchevaluatingllmagent} during the generation process, which significantly multiply the computational overhead. Finally, considering data privacy issues, local deployment of small-scale LLMs with strong performance and reducing training costs is important, including the volume and quality of training data. These limitations pose significant barriers to deploying LLMs with strong optimization-solving capabilities in small and medium-sized enterprises and even on mobile devices.

To address these challenges, we propose \textsc{MiniOpt}, an RLVR-based methodology designed to enhance the {optimization generalization} of small-scale LLMs under limited data and training resources. We systematically design the training process, training data, and algorithms of the model, introducing the \textit{reasoning-to-model-and-solve} paradigm. Specifically, we adapt the classic ``model and solve'' paradigm of LLMs for optimization problems into a verifiable Chain-of-Thought (CoT) for Reinforcement Learning (RL), thereby facilitating effective RLVR training. 
\textbf{To efficiently utilize limited data, we propose a two-stage RL framework that divides the training into progressive stages aligned with the evolution of model capabilities}. Initially, we perform Mid-Training on a small subset of annotated data to prevent training collapse induced by reward sparsity. Subsequently, the first RL stage employs a training set of lower difficulty to enable the policy model to master the \textit{reasoning-to-model-and-solve} paradigm. The second stage then further enhances performance and optimization generalization on more challenging datasets. 
Addressing the scarcity of annotated CoT data and the associated verification challenges, MiniOpt employs five-element tuple modeling as a verifiable structural abstraction of the CoT within the RL framework. This approach mitigates verification complexity while simultaneously enhancing result accuracy and model generalization. Furthermore, we design a \textbf{structured reward function, \textit{OptReward}}, which facilitates robust RLVR. Specifically, it incorporates a modeling completeness verification to circumvent the need for content verification of the optimization modeling, while simultaneously constraining the RL reasoning trajectory in the vicinity of the correct modeling formulation to mitigate reward hacking. 
Building upon this systematic architectural design, we also introduce tailored modifications to the GPRO algorithm, thereby fully leveraging limited samples to conduct highly efficient training and achieve robust optimization generalization. Ultimately, \textbf{the aforementioned synergistic design of the RLVR framework significantly reduces the reliance on annotated data, as well as the overall training and inference costs}.

Building upon the aforementioned methodology, this paper conducts extensive experiments with MiniOpt-3B and -8B on 8 benchmarks across different optimization types and problem scenarios. The results demonstrate its strong optimization generalization. \textbf{Compared with baselines with fewer than 10B parameters, MiniOpt-3B achieves the best performance}. When evaluated against baselines exceeding 10B parameters, MiniOpt-3B remains competitive compared to LLMOPT-14B, and surpasses the general thinking model GPT-5 by 2.11 percentage points. MiniOpt-3B achieves a competitive average SA compared with DeepSeek-R1 while consuming only about 24.40\% of the output tokens used by DeepSeek-R1 across benchmarks of varying difficulty. Notably, \textbf{MiniOpt lies on the empirical Pareto frontier of both parameter scale and the SA metric}. Furthermore, results from ablation studies and discussion indicate that applying RLVR within the \textit{reasoning-to-model-and-solve} paradigm yields substantial performance gains. MiniOpt demonstrates advantages across all eight benchmarks and parameter-decoupled settings.

Beyond evaluating end-to-end optimization performance, we further conduct a complementary mechanistic analysis to better understand where optimization generalization originates. we construct a parameter-decoupled evaluation benchmark, \text{OptiBench-Separate}, by externalizing the tabular parameters in OptiBench into multiple CSV files while preserving exactly the same optimization structures and problem semantics. \textbf{Experimental results indicate that both general-purpose large language models and specialized optimization models exhibit noticeahle performance degradation under this parameter-decoupled setting}, while MiniOpt demonstrates substantially less susceptibility to such degradation, suggesting its strong generalization capability in optimization tasks. Although models often correctly generate data-loading operations (e.g., \texttt{pd.read\_csv}) and recognize that parameters should be obtained from external files, they frequently hallucinate numerical coefficients when formulating the optimization model instead of consistently grounding them in the loaded data. The main contributions of this paper are summarized as follows:
\begin{enumerate}
    \item We propose MiniOpt, a two-stage RL framework that enables small-scale LLMs to achieve strong optimization generalization under limited training resources through the proposed reasoning-to-model-and-solve paradigm.
    \item We introduce OptReward and OptGRPO, an optimization-oriented reward function and policy optimization algorithm that together enable efficient RL training without requiring expert reasoning annotations or expensive intermediate verification.
    \item Extensive experiments across multiple optimization benchmarks demonstrate that MiniOpt establishes a favorable Pareto frontier between model scale and optimization performance, while consistently outperforming existing learning-based methods under both conventional and parameter-decoupled evaluation protocols.
    \item We revealed the bottleneck of learning-based methods in optimizing generalization through parameter-decoupling evaluation.
\end{enumerate}

The remainder of this paper is organized as follows. Section~\ref{related_work} reviews related work. Section~\ref{methodology} presents the proposed MiniOpt framework for learning optimization generalization under limited resources. Section~\ref{sec:exp} reports extensive experiments evaluating both the optimization generalization capability of MiniOpt and the proposed parameter-decoupled evaluation protocol. Finally, Section~\ref{discussion} discusses additional analyses before concluding the paper.

\section{Related Work}
\label{related_work}
\textbf{LLMs for Modeling and Solving Optimization Problems.} For modeling and solving optimization problems with LLMs, there are already a variety of benchmarks\cite{huang2025llmsmathematicalmodelingbridging, pmlr-v235-ahmaditeshnizi24a, yang2024benchmarking}. Challenging benchmarks like Mamo \cite{huang2025llmsmathematicalmodelingbridging} and OptiBench \cite{yang2025optibench} have led to numerous studies utilizing LLMs to solve optimization problems. Prompt-based approaches such as OptiMUS\cite{pmlr-v235-ahmaditeshnizi24a}, CoE\cite{xiao2024chainofexperts}, OptiTree\cite{liu2025optitree} and LEAN-LLM-OPT\cite{leanllmopt} utilize the powerful generation capability of LLMs to generate the solver code of the optimization problem through multi-stage pipeline, without performing any post-training. Learning-based methods enhance LLMs' capabilities in modeling and solving mathematical problems. For example, LLaMoCo\cite{ma2024llamocoinstructiontuninglarge} proposes an SFT-based framework comprising a meticulously designed instruction set and a two-stage training methodology that incorporates contrastive learning warm-up followed by SFT. LLMOPT \cite{JiangShu2025llmopt} and NER4OPT\cite{Kadioglu24, singirikonda2025text2zinccrossdomaindatasetmodeling} adopt a two-stage training process of modeling the optimization problems first and then solving them by generating solution code.
StepORLM\cite{zhou2025steporlmselfevolvingframeworkgenerative} synthesizes a large volume of process data to simultaneously learn a policy model and a Process-Reward-Model (PRM), employing alternating training to produce a modeling-and-solving model with reliable process consistency. Similarly, BPP-Search\cite{wang-etal-2025-bpp} generates process data and trains a PRM alongside a preference selection model under a Tree-of-Thought search paradigm, thereby simulating reasoning processes to derive robust results.

\textbf{Reinforcement Learning with Verifiable Reward.} While Reinforcement Learning from Human Feedback (RLHF)\cite{10.5555/3600270.3602281} plays a crucial role in post-training alignment, it suffers from high annotation costs and inherent human bias\cite{xiao2025surveyoptimizationmodelingmeets}.
Reinforcement Learning with Verifiable Reward (RLVR) \cite{lambert2024tulu} leverages externally grounded, easily verifiable rewards (e.g., rule-based reward) to provide dense and structurally simple supervision\cite{wang-etal-2024-math, xie2025logicrlunleashingllmreasoning, gao2024designingeffectiverlreward}. Its practicality is especially valuable in real-world black-box systems\cite{zhang-etal-2025-codedpo, xin2025deepseekproverv, pan2025training}, where verification is typically feasible only at the output stage, making RLVR a broadly applicable paradigm for aligning LLMs. For example, SIRL\cite{chen2025solverinformedrlgroundinglarge} and OR-R1\cite{ding2025orr1automatingmodelingsolving} significantly improve the model's performance through the RLVR training paradigm.

\section{Methodology: The Proposed MiniOpt}
\label{methodology}

\begin{figure*}[t]
  \centering
    \includegraphics[width=0.90\textwidth]{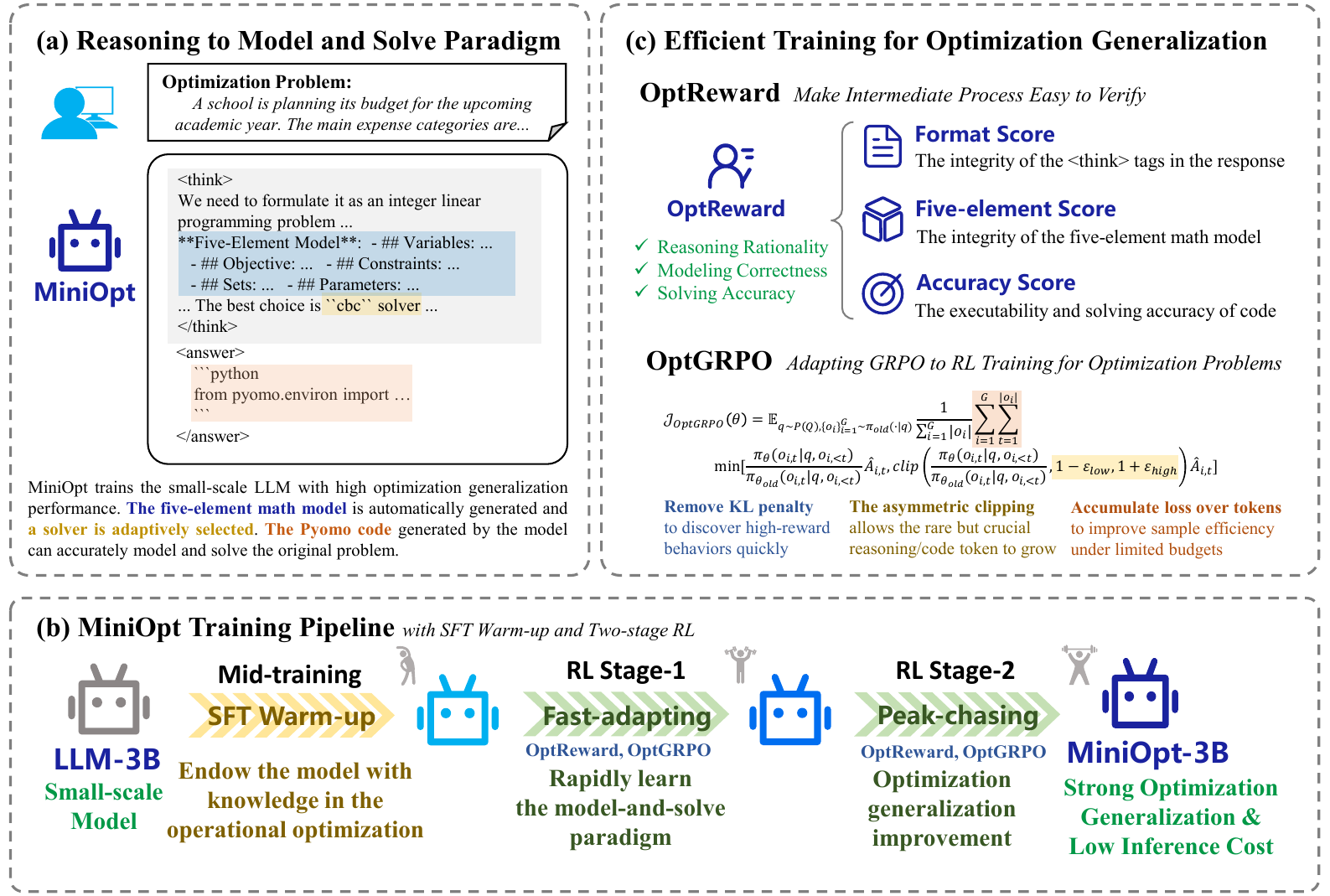}
    \caption{An overview of the proposed MiniOpt training paradigm. Sub-figure (a) demonstrates the reasoning-to-model-and-solve paradigm of MiniOpt, encompassing problem modeling and solver adaptation during the thinking (i.e., reasoning) process of RL, and solution code generation in the response. Sub-figure (b) illustrates the training pipeline of MiniOpt, which involves sequential execution of SFT (mid-training) as warm-up followed by two-stage RL (post-training). Sub-figure (c) presents the reward function OptReward and training algorithm OptGRPO used in MiniOpt's RL training.}
    \label{fig:miniopt_r}
\end{figure*}

\subsection{Overview}
\label{sec:method_overview}
This paper studies how to endow small-scale LLMs with strong optimization generalization under tight data and compute budgets. We introduce MiniOpt, a reasoning-driven training paradigm for modeling and solving optimization problems, whose training framework is shown in Figure~\ref{fig:miniopt_r}. MiniOpt integrates a five-element tuple structure into the Chain-of-Thought (CoT) and leverages a structured \textit{OptReward} to achieve RLVR training with strong optimization generalization. Through this framework, MiniOpt formulates the pipeline from natural language problems to executable solver code as a single, end-to-end verifiable task.

\subsection{Reasoning-to-Model-and-Solve Paradigm}
\label{sec:method_paradigm}
The optimization problem can be described as the following formulation:
\begin{equation}
\label{formulation:opt}
    \min_{\x \in \X \subseteq \mathbb{R}^D} f(\x) \, , \,\,\,\,
    {\rm s.t. } \,\, G(\x) \leq \boldsymbol{c} \,\, ,
\end{equation}
where $\x \in \X \subseteq \mathbb{R}^D$ denotes the $D$-dimensional decision \text{variable}, and $I=\{1,2,\dots, D\}$ is the index \text{set}. $\X$ is the feasible region and $\boldsymbol{c} \in \mathbb{R}^m$ provides the upper bounds. The constants in \text{objective function} $f(x)$ and \text{constraints} $G(x)$ form the \text{parameter} set, which also includes $c$.
The five-element optimization formulation, $\mathcal{M} = $(\text{Variables}, \text{Objective}, \text{Constraints}, \text{Sets}, \text{Parameters}) maps one-to-one to the components of an optimization problem.

As shown in subfigure (a) in Figure~\ref{fig:miniopt_r}, we introduce a \textit{reasoning-to-model-and-solve} paradigm that turns a natural language optimization problem into a single verifiable objective. The paradigm is enforced by two compulsory and parsable segments \texttt{<think>} $\dots$ \texttt{</think>} and \texttt{<answer>} $\dots$ \texttt{</answer>}. 

The first segment, enclosed by \texttt{<think>} $\dots$ \texttt{</think>}, contains all modeling-related contents. It analyzes the problem statement, specifies a verifiable \text{five-element optimization formulation}, and determines an appropriate open-source solver. Since optimization structures are largely invariant across different problem instances of the same type of optimization problem, while numerical parameters may vary significantly, learning optimization structures provides substantially stronger generalization than directly fitting problem-specific parameter configurations.

During the reasoning process, to enhance the capability of LLM for solving optimization problems across diverse optimization types, it analyzes the problem during the reasoning phase and selects different open-source solvers for the given problem depending on the optimization types and the characteristics of the solvers, thereby improving the match between the problem and the selected solver.

The second segment, enclosed by \texttt{<answer>} $\dots$ \texttt{</answer>}, converts the five-element formulation into an executable Pyomo program that models the problem, invokes the solver, solves the instance, and prints the numerical answer. By constraining responses in this manner, we collapse the action space from free-form language to a programmable artifact whose intermediate structure and final result can be deterministically parsed and verified. The \texttt{<answer>} segment must implement the blueprint as a single Python code fence that contains a complete Pyomo script. Because these outputs are easy to parse and verify, OptReward in Section~\ref{sec:method_opt_reward} can automatically score the response format, the five-element formulation and the numerical accuracy based on rules through an automated procedure, providing low-cost supervision. The prompt used for solver selection and the reasoning-to-model-and-solve training are provided in the supplementary material.

\subsection{Training Pipeline of MiniOpt}
\label{sec:method_training_pipeline}

Based on the paradigm described in Section \ref{sec:method_paradigm} and the key techniques introduced in Section \ref{keycomps}, we propose a training pipeline that enables MiniOpt to learn from limited resource constraints while achieving strong optimization generalization. This pipeline begins with a lightweight SFT as mid-training followed by a two-stage RL under OptReward. The illustration of the training pipeline is shown in subfigure (b) of Figure~\ref{fig:miniopt_r}.

\subsubsection{Mid-Training based on SFT}
\label{sec:method_sft}
LLMs acquire broad knowledge through large-scale general pretraining. However, for specialized domains, small-scale models often struggle to cultivate the necessary capabilities effectively. Mid-training serves as a bridge between pre-training and post-training, strengthening domain-specific capabilities through targeted training data\cite{tu2025surveyllmmidtraining}. In the proposed training pipeline, the mid-training phase employs a small-scale and diverse dataset from the operational optimization field to conduct SFT on the base model, thereby providing effective modeling and solving trajectories for the subsequent two-stage RL training.

This SFT warm-up establishes a starting point upon which our two-stage RL can subsequently focus on paradigm acquisition and optimization generalization, it does not adopt the reasoning-to-model-and-solve paradigm introduced in Section~\ref{sec:method_paradigm}.
Based on the annotations of optimization types and problem-scenario tags assigned to each sample, the training data for mid-training are constructed from a small-scale subset of OptMath-Train-Pyomo. Within this subset, we ensure an equal number of linear and nonlinear problems. Since nonlinear problems are scarce in OPTMath-Train, we supplement the shortfall with MILP problems in the dataset to reach the desired equal distribution. Additionally, all samples labelled as "unknown" are included to preserve both the difficulty and diversity of the data source during the warm-up training phase. The processing pipeline of the dataset for this SFT training and the prompt used for data annotation are provided in the supplementary material.

\subsubsection{The Two-Stage RL}
\label{sec:method_two_stage_rl}
After the SFT phase, the models can undergo RL training more effectively. We employ a two-stage RL training under the same reward function and policy optimization algorithm for paradigm acquisition and optimization generalization, respectively. Both stages share the pipeline of parsing, executing, and scoring in Section~\ref{sec:method_paradigm}, they differ in training data. Algorithmically.

Stage-1 aims to enable the model to acquire the reasoning-to-model-and-solve paradigm: the model must generate a valid \texttt{<think>} / \texttt{<answer>} pair, produce the executable Pyomo code, and make a coherent solver choice. To this end we train on 1,585 relatively easy problems, so that most signal arises from the formatting and structural components of OptReward, rapidly improving executability and the ability to solve the optimization problem of natural language description.

Stage-2 focuses on optimization generalization once the paradigm is established. The training distribution shifts to the problems with diverse optimization types and problem scenarios, and the optimization emphasis moves to the accuracy score, encouraging refined modeling and solving behaviors (e.g., variable/constraint formulation and solver selection). We sample a data subset from the candidate pool mentioned in Section~\ref{sec:method_sft} subject to two constraints: (i) type-uniform coverage, with the same number of instances per type, and (ii) within each type, the scenario frequencies match the distribution in the full dataset. The resulting data serves as the Stage-2 RL training set. The dataset for the first stage is the union of the NL4Opt\cite{pmlr-v220-ramamonjison23a} and ICML Competition\cite{icmlcompetetion2024} training splits. Detailed information on the construction of the training set is provided in the supplementary material.

Such a training strategy makes efficient use of limited data and reduces training costs, and ultimately allows for strong optimization generalization of LLMs with even small parameters and limited computational resources.

\subsection{Efficient Training for Optimization Generalization}
\label{keycomps}
Building upon the reasoning-to-model-and-solve paradigm in Section \ref{sec:method_paradigm}, we propose two key components for the RL training of MiniOpt as shown in subfigure (c) in Figure \ref{fig:miniopt_r}: An easily verifiable reward function OptReward, and an improved algorithm OptGRPO builds upon GRPO\cite{shao2024deepseekmath}.

\subsubsection{OptReward: Verifiable Rewards Designed for MiniOpt}
\label{sec:method_opt_reward}

Building upon the reasoning-to-model-and-solve pipeline, we propose \textbf{OptReward} as the core mechanism for RLVR training, serving not only as a verification tool for optimization modeling but also as a gating mechanism to control the stages of RL training. It comprises three automatically computed components: format correctness, structural completeness, and numerical accuracy. Each component is derived from the deterministic parsing or execution of the model's output, which not only ensures the completeness of the problem modeling but also enables verification to scale with computational time, substantially reducing the overall verification cost.

\textbf{Format Score:} The format score $S_{\mathrm{fmt}}$ validates the response format. A response must contain exactly one \texttt{<think>}…\texttt{</think>} and one \texttt{<answer>}$\dots$\texttt{</answer>} in the correct order. If all conditions hold, we assign $S_{\mathrm{fmt}}=+1$, otherwise $S_{\mathrm{fmt}}=-1$. If the specified format is not present in the response, we deterministically set the remaining components to their error defaults, $S_{\mathrm{five}}=-1$ and $S_{\mathrm{acc}}=-2$, so that the total reward immediately reaches the global minimum. This forces the model to adopt the correct response format early in training and prevents expensive evaluation of malformed samples.

\textbf{Five-element Score:}
To avoid ground-truth labeling of five-element content and the bias it may introduce, we use a presence-based rule aligned with the paradigm in Section~\ref{sec:method_paradigm}.
Conditional on valid formatting, we compute the five-element score $S_{\mathrm{five}}$. In the \texttt{<think>} segment, the model response is expected to include five labelled summaries starting with ``\#\# Sets:'', ``\#\# Parameters:'', ``\#\# Variables:'', ``\#\# Objective:'', and ``\#\# Constraints:''. Each present summary contributes 0.2 points to the score; if none is present, we assign $S_{\mathrm{five}}=-1$. This structure shaping keeps the modeling blueprint parsable.
\begin{equation}
S_{\mathrm{five}} =
\begin{cases}
0.2 \displaystyle\sum_{k=1}^{5} I_k, & \text{if } \displaystyle\sum_{k=1}^{5} I_k \ge 1\,, \\[4pt]
-1, & \text{if } \displaystyle\sum_{k=1}^{5} I_k = 0\,,
\end{cases}
\end{equation}
where $I_k = 1$ if the $k$-th required element $e_k$ is present in the \texttt{<think>} segment, and $I_k = 0$ otherwise.

\textbf{Accuracy Score:} 
The accuracy score is obtained by executing the Pyomo program contained in \texttt{<answer>}. If the program cannot be extracted or execution fails, we assign $S_{\mathrm{acc}} = -2$. When execution succeeds, we retrieve the optimal objective value $\hat{f}$ from the model output and compare it with the ground-truth value $f^\star$; if they are equal, we assign $S_{\mathrm{acc}} = 2$, otherwise $S_{\mathrm{acc}} = -1.5$. 
\begin{equation}
S_{\mathrm{acc}} =
\begin{cases}
+2, & \text{if execution succeeds and } \hat{f} = f^\star, \\[4pt]
-1.5, & \text{if execution succeeds but } \hat{f} \neq f^\star, \\[4pt]
-2, & \text{if no executable code or execution fails}\,. \\
\end{cases}
\end{equation}
Combining the components with the formatting gate yields the total OptReward as follows:
\begin{equation}
R =
\begin{cases}
-4, & \text{if}\ S_{\mathrm{fmt}} = -1, \\[4pt]
S_{\mathrm{fmt}} + S_{\mathrm{five}} + S_{\mathrm{acc}}, & \text{if}\ S_{\mathrm{fmt}} = 1.
\end{cases}
\end{equation}
\textbf{The proposed reward primarily evaluates whether the generated optimization model correctly represents the intended optimization structure, rather than rewarding memorization of instance-specific numerical parameters. This design is consistent with our objective of learning optimization generalization through structural reasoning.} By OptReward, the format score enforces the strict \texttt{<think>} / \texttt{<answer>} format, the five-element score shapes a complete modeling blueprint in the think phase, and the accuracy score certifies correctness through equality of optimal objective values, enabling low-cost verifiable RL for problems in the field of optimization.

\subsubsection{OptGRPO: Training Small-Scale LLMs with Limited Resources}
\label{sec:modified_grpo}

Considering the specific requirements of the task, we build upon the reasoning-to-model-and-solve paradigm and OptReward to refine the GRPO algorithm, introducing the \textbf{OptGRPO}.
In the second stage of RL, MiniOpt is dedicated to exploring complex problem-solving patterns. In this context, a symmetrically designed upper bound for importance weight clipping can induce premature convergence, thereby hindering the model's ability to discriminate between correct and incorrect solutions. Similarly, the standard KL divergence penalty inherently restricts the exploration capacity of the policy model. 
To address these limitations, we set the coefficient on the KL penalty $\beta=0$ to remove the KL penalty to encourage greater exploration by the model, and replace symmetric clipping with an asymmetric interval $[1-\varepsilon_{\text{low}},\, 1+\varepsilon_{\text{high}}]$ with a higher upper clipping threshold $\varepsilon_{\text{high}}$ than the lower threshold $\varepsilon_{\text{low}}$, to encourage the exploration of low-probability (yet potentially optimal) patterns. This relaxes the trust region on probability increases while keeping a firm lower bound on decreases, which empirically improves executability in stage-1 RL training and supports generalization in stage-2 RL training. Specifically, we raise $\varepsilon_{\text{high}}$ to 0.28 during the training. Although a higher clipping threshold may potentially introduce high variance, training stability is empirically guaranteed because the modeling format of the rollouts is already strictly constrained by the five-element tuple structure within the CoT. Furthermore, we adopt a token-level policy gradient loss computation. This design amplifies the reward signals for valid solution patterns, ensuring that high-quality reasoning trajectories are not disproportionately penalized or underestimated, thereby providing a more precise and granular reward signal for the target solving behaviors. Ultimately, through OptGRPO, the \textit{reasoning-to-model-and-solve} training paradigm is executed with significantly enhanced efficiency and efficacy. The final loss of OptGRPO is as follows:
\begin{align}
\mathcal{J}_{\text{OptGRPO}}(\theta) 
&=
\mathbb{E}_{q \sim P(Q), \{o_i\}_{i=1}^G \sim \pi_{\theta_{\text{old}}}(\cdot|q)}
\frac{1}{\sum_{i=1}^G |o_i|} \sum_{i=1}^{G} \sum_{t=1}^{|o_i|}  \notag \\ 
& \qquad \min [ \rho_{\theta} \hat{A}_{i,t}, \text{clip}(\rho_{\theta}, 1-\textcolor{black}{\varepsilon_{\text{low}}}, 1+\varepsilon_{\text{high}})\hat{A}_{i,t} ]\,,
\end{align}
\begin{align}
\rho_{\theta}=\frac{\pi_\theta(o_{i,t}|q, o_{i,<t})}{\pi_{\theta_{\text{old}}}(o_{i,t}|q, o_{i,<t})}\,.
\end{align}
\subsection{Structure-Oriented Optimization Learning}
Although optimization problems contain both optimization structures and instance-specific numerical parameters, the objective of MiniOpt is to learn optimization structural understanding, which are invariant across different problem instances and constitute the essence of optimization generalization. The proposed reasoning-to-model-and-solve paradigm therefore constitutes a structure-oriented learning paradigm. During reinforcement learning, the policy is optimized by verifying whether the generated optimization formulation correctly captures the intended optimization structure and whether the resulting optimization solution is valid through RLVR. Consequently, the reinforcement signal primarily reflects the correctness of structural reasoning instead of memorization of instance-specific numerical values. This design enables MiniOpt to improve optimization generalization while maintaining stable reinforcement learning and avoiding expensive intermediate supervision. This learning paradigm is fundamentally different from conventional supervised fine-tuning~\cite{JiangShu2025llmopt, Huang_2025}, which directly learns mappings from problem descriptions to solver codes and may implicitly fit both optimization structures and instance-specific numerical patterns.

\section{Experiments}
\label{sec:exp}
We evaluate MiniOpt models on diverse optimization benchmarks spanning multiple types and scenarios to assess whether small-scale parameter LLMs (3B/7B) can achieve strong optimization generalization ability. 

To evaluate solution correctness, we report Solving Accuracy (SA) as the primary metric, SA measures the proportion of generated code samples whose execution results match the ground-truth optimal solutions provided by the benchmarks. A solution is deemed correct if the returned optimal objective value matches the reference solution within a tolerance of $10^{-1}$; otherwise, it is considered unsuccessful. Comparisons cover general LLMs, general reasoning LLMs, prompting-based baselines and learning-based baselines. The experiments are designed to answer the following five research questions.

\textbf{(Q1) Optimization Generalization Ability at Small-Scale LLMs.} To what extent can MiniOpt at 3B/7B achieve high SA and ER across types and scenarios, and how does it compare with larger reasoning LLMs and prior learning-based approaches?

\textbf{(Q2) Pareto Front of Performance vs. Cost.} What is the limit of the scale of model parameters for achieving strong optimization generalization?

\textbf{(Q3) Importance of Each Stage in the Training Pipeline of MiniOpt.} How critical are the lightweight SFT warm-up and the two-stage RL to the performance of MiniOpt?

\textbf{(Q4) Importance of the OptReward and OptGRPO.} How does the proposed OptReward and OptGRPO in MiniOpt contribute to boost SA and ER in modeling and solving optimization problems?

\textbf{(Q5) Mechanistic Analysis of Optimization Generalization.}
We introduce a parameter-decoupled evaluation protocol that disentangles the understanding of the optimization structure from the extraction of numerical parameters, providing complementary evidence for the effectiveness of MiniOpt.

While Q1--Q4 focus on how optimization generalization can be learned effectively under limited resources, Q5 investigates where does MiniOpt's improvement come from? On the basis of verifying the effectiveness of MiniOpt in the standard benchmark, we further design a complementary diagnostic evaluation to explain whether MiniOpt's optimization generalization ability comes from understanding the optimization structure or from its parameter processing ability.

\subsection{Experimental Setup}
\label{sec:exp_setup}

Since widely used packages such as Gurobi and COPT are closed-source, generating code for them may entail licensing costs, and a model’s generalization ability across different modeling languages and solvers depends on the proportion of relevant data in its pretraining and post-training corpora. Therefore, we adopted the open-source, solver-agnostic Pyomo modeling language as the language for our training and inference solver code. As for the solvers employed, the key consideration lies in their ability to select the appropriate solver for a specific optimization type. Therefore, this paper selects four types of solvers to cover the solving requirements of as many optimization problem types as possible, so as to automatically adapt to the problem types when generating the solving code. The training configuration of MiniOpt is as follows: SFT is conducted for 4 epochs; the first stage of RL involves 45 training steps, while the second stage consists of 25 steps. During inference, a temperature of 0.7 is applied. For additional experimental settings, please refer to the anonymous repository mentioned in the abstract of this paper.

The evaluation encompasses eight benchmarks about operational optimization: NL4Opt \cite{pmlr-v220-ramamonjison23a}, Mamo (Easy and Complex subsets, abbreviated as Mamo.E and Mamo.C, respectively) \cite{huang2025llmsmathematicalmodelingbridging}, IndustryOR\cite{Huang_2025}, NLP4LP\cite{pmlr-v235-ahmaditeshnizi24a}, ComplexOR\cite{xiao2024chainofexperts}, OptiBench\cite{yang2025optibench}, and ICML Competition (ICML.C)\cite{icmlcompetetion2024}. We follow the same setting in LLMOPT\cite{JiangShu2025llmopt} to ensure consistency and comparability. For the newly included dataset, OptiBench and ICML Competition, we adhere to their original data splits provided by the authors.

\subsection{Analysis of Optimization Generalization}
In this section, we compare MiniOpt with general LLMs (Qwen2.5, DeepSeek-V3), general thinking LLMs (Qwen3, DeepSeek-R1, Gemini-2.5-Pro, GPT5), prompt-based methods (Chain of Experts, OptiMUS, Reflexion), learning-based methods (OptMATH-7B, LLMOPT-14B, Step-OPT-7B/3B), demonstrating the optimization generalization capability of MiniOpt. Table~\ref{tab:opti_general_sa_ranked} summarizes SA on eight benchmarks that span 7 optimization types and 22 scenarios. The information on these methods and the statistics on problem categories and scenarios of benchmarks are provided in supplementary material, as well as the \textit{Executable Rate} metrics for all the methods on the 8 benchmarks and its analysis.

\begin{table*}[!t]
  \centering
  \scriptsize
  \setlength{\tabcolsep}{3pt}
  \caption{Comparison of the SA metric across 8 benchmarks with rankings (NL4Opt, ICML Competition, Mamo Easy, Mamo Complex, NLP4LP, ComplexOR, IndustryOR, OptiBench). \textbf{Bold} indicates 1st, \uwave{wavy underline} indicates 2nd, \underline{underline} indicates 3rd. ``Rank$^{*}$'' represents the result of sorting methods among parameter sizes below 10B.}
  \label{tab:opti_general_sa_ranked}
  \resizebox{0.99\linewidth}{!}{
  \begin{tabular}{l | l | c |c| c| c c c c c c c c}
    \toprule
    Category & Models / Methods & \textbf{Avg.} & \textbf{Rank} & \textbf{Rank$^{*}$} & \textbf{NL4Opt} & \textbf{ICML.C} & \textbf{Mamo.E} & \textbf{Mamo.C} & \textbf{NLP4LP} & \textbf{Com.OR} & \textbf{Indus.OR} & \textbf{OptiBench}\\
    \midrule
    \multirow{4}{*}{\makecell{General Models}} 
       & Qwen2.5-3B-Instruct & 11.23 & 18  & 9 & 19.13 & 18.78 & 17.18 & 2.37 & 18.60 & 0.00 & 2.00 & 11.74 \\
       & Qwen2.5-7B-Instruct & 33.20 & 14 & 6 & 53.48 & 51.71 & 35.58 & 4.27 & 55.37 & 16.67 & 13.00 & 35.54 \\
        & Qwen2.5-14B-Instruct & 47.46 & 10 & - & 67.39 & 63.17 & 80.21 & 15.64 & 67.36 & 22.22 & 22.00 & 41.65 \\
        & DeepSeek-V3\quad(671B) & \underline{60.14} & {3} & -  & 78.26 & \underline{77.56} & \underline{84.82} & 26.54 & \textbf{79.34} & \underline{44.44} & 26.00 & \underline{64.13} \\

    \midrule
    \multirow{6}{*}{\makecell{General Models \\(Thinking)}}
     & Qwen3-4B &  11.16 & 19 & 8 & 16.52 & 17.56 & 13.80 & 6.64& 15.29 & 5.56 & 2.00 & 11.90  \\
      & Qwen3-8B & 21.79 & 16 & 7  & 30.87 & 29.51 & 23.93 & 9.95 & 34.71 & 11.11 & 6.00 & 28.26 \\
      & Qwen3-14B & 23.78 & 15  & - & 24.35 & 22.68 & 36.20 & 11.37 & 19.42 & 38.89 & 15.00 & 22.31 \\
      & DeepSeek-R1\quad(671B) & \uwave{60.85} & {2} & -  & \underline{83.91} & 75.37 & 74.54 & {39.81} & 69.83 & \underline{44.44} & \textbf{32.00} & \textbf{66.94} \\
      & Gemini-2.5-Pro & 57.39 & 7 & -  & 78.26 & 71.22 & 65.95 & 30.81 & 73.55 & \uwave{50.00} & \underline{28.00} & 61.32 \\
      & GPT-5 & 57.54 & 6 & -  & 80.43 & 73.66 & 58.12 & 23.22 & 73.14 & \textbf{61.11} & 26.00 & \uwave{64.63} \\

    \midrule
    \multirow{3}{*}{\makecell{Prompt-based \\Methods}}
    
    & Chain-of-Experts & 45.78 & 11  & - & 66.52 & 56.59 & 63.65 & 22.75 & 59.09 & 33.33 & 19.00 & 45.29 \\
    & OptiMUS & 20.65 & 17 & - & 13.48 & 33.17 & 37.27 & 11.85 & 18.18 & 16.67 & 8.00 & 26.61 \\
    & Reflexion & 45.54 & 12 & - & 56.52 & 52.20 & \underline{84.82} & 18.01 & 53.72 & 38.89 & 19.00 & 41.16 \\

    \midrule
    \multirow{2}{*}{\makecell{\\Learning-based \\Models}}
    & Step-OPT-Qwen2.5-3B & 39.76 & 13 & 5 & 41.30 & 38.54 & 75.31 & 20.85 & 53.31 & 27.78 & 21.00 & 40.00 \\    
    & Step-OPT-Qwen2.5-7B & 52.22 & 9 & 4 & 77.83 & 57.32 & 69.33 & \uwave{50.24} & 48.35 & 38.89 & 27.00 & 48.76 \\
    & OptMATH-7B & 54.62 & 8 & {3} & 78.70 & 66.83 & 84.20 & 34.12 & 68.60 & 33.33 & 19.00 & 52.23 \\
    & LLMOPT-14B & 60.10 & 4 & -  & 80.28 & 75.35 & \uwave{89.53} & 44.08 & 73.42 & 35.29 & \uwave{29.00} & 53.83 \\
    
    \midrule
    \multirow{3}{*}{\makecell{\textbf{\qquad Ours}}}
    
    & MiniOpt-3B & 59.65 & 5 & {2} & \uwave{93.04} & \uwave{78.05} & 84.36 & \underline{47.39} & \underline{74.38} &  27.78& 17.00 & 55.21 \\
    & MiniOpt-7B & \textbf{64.76} & {1} & {1} & \textbf{95.22} & \textbf{83.41} & \textbf{89.57} & \textbf{51.18} & \uwave{74.79} & {38.89} & 25.00 & 60.00 \\
      
    \bottomrule
  \end{tabular}
  }
\end{table*}

\textbf{Overall Performance (Answer to Q1).} 
Across all eight benchmarks, MiniOpt-7B achieves the strongest average performance among all baselines. Notably, MiniOpt-3B surpasses all prompt-based and general methods. For example, compared to Gemini-2.5-Pro, the state-of-the-art general thinking model, MiniOpt-3B achieved an average SA of 2.26\% higher. Compared to DeepSeek-V3 and DeepSeek-R1, MiniOpt-3B scored 0.49\% and 1.20\% lower, respectively. Furthermore, MiniOpt-7B achieves the highest average SA of 64.76\%, significantly raising the performance ceiling of MiniOpt in practical applications.

\textbf{Competitiveness of Small-Scale Models (Answer to Q1).}
MiniOpt remains competitive even at smaller scales. MiniOpt-3B reaches an average SA of 59.65\%, which already matches several much larger reasoning models (e.g., the average SA higher than GPT-5 at 57.54\%) and clearly outperforming general-purpose 14B pre-trained models (e.g., +12.19\% over Qwen2.5-14B-Instruct on average). Performance grows smoothly with scale under the same training pipeline, the average SA increases by +5.11\% when the model size grows from 3B to 7B.

\textbf{Challenging Benchmarks (Answer to Q1).}
On the challenging benchmark Mamo.Complex, MiniOpt-3B and -7B achieve 47.39\% and 51.18\% SA, respectively, outperforming all the general models and general thinking models. Even on extremely high-dimensional test sets such as Indus.OR, where the smaller MiniOpt models do not achieve the highest SA because of their limited instruction-following capability, it still delivers competitive performance levels in both metrics.

\textbf{Breadth across Types and Scenarios (Answer to Q1).}
To rigorously evaluate optimization generalization, we benchmark MiniOpt-3B and MiniOpt-7B across three difficulty tiers: simple (ICML.C), medium (OptiBench), and hard (Mamo.C). The evaluation encompasses diverse application scenarios (Supply Chain, Transportation, Resource) and problem formulations (LP, IP, MILP). The results highlight a clear scaling law within our framework. While MiniOpt-3B demonstrates competent baseline performance on simpler tasks (e.g., 94\% in Transportation and 93\% in Resource on ICML.C), MiniOpt-7B consistently bridges the gap on complex reasoning tasks. On the ICML.C, the 7B variant elevates MILP accuracy from 31\% to 50\%. This scaling advantage persists under increased difficulty; on the hard-difficulty Mamo.C benchmark, MiniOpt-7B maintains a 10\% to 13\% absolute improvement over the 3B model in LP, IP, and MILP formulations. Furthermore, both models exhibit remarkable robustness in Resource scenarios, with the 7B model sustaining a 75\% success rate even on the most complex tasks. These findings validate that MiniOpt provides a scalable and reliable foundation for LLM-based optimization modeling.

\begin{table*}[h]
  \centering
  \scriptsize
  \setlength{\tabcolsep}{3pt}
  \caption{Ablation study (MiniOpt-3B) on the SA metric across 8 benchmarks. Herein, ``w'' denotes ``with'' and ``w/o'' denotes ``without''.}
  \label{tab:ablation_sa}
  \resizebox{0.99\linewidth}{!}{
  \begin{tabular}{l l | c | c c c c c c c c}
    \toprule
    Category & Model / Method & \textbf{Avg.} & \textbf{NL4Opt} & \textbf{ICML.C} & \textbf{Mamo.E} & \textbf{Mamo.C} & \textbf{NLP4LP} & \textbf{Com.OR} & \textbf{Indus.OR} & \textbf{OptiBench} \\
    \midrule
    \multirow{8}{*}{\textbf{Ablations}}
        & MiniOpt-3B                     & \textbf{59.65} & {93.04} & {78.05} & 84.36 & {47.39} & {74.38} &  {27.78}& 17.00 & 55.21 \\
        
        & MiniOpt-3B w/o SFT Warm-up     & 56.60 & 89.13 & 76.83 & 84.36 & 28.44 & 72.73 & {27.78} & {19.00} & 54.55 \\     
        
        & MiniOpt-3B w/o RL              & 27.24 & 40.00 & 35.37 &50.15& 3.79 & 41.74 & 11.11  & 7.00   & 28.76  \\     
        
        & MiniOpt-3B w/o Two-stage RL    & 55.07 & 89.13 & 73.41 & 79.75 & 33.18 & {74.79} & 22.22 & 16.00 & 52.07 \\     
        
        & MiniOpt-3B w/ Random Selection & 58.33 & 93.04 & 79.02 & 82.06 & 48.34 & 73.97 & 16.67 & 17.00 & {56.53} \\     
        
        & MiniOpt-3B w/o OptReward       & 57.96 & 89.13 & 76.34 & 83.90 & 38.86 & 72.73 & 33.33 & 16.00 & 53.39 \\    
        
        & MiniOpt-3B w/ GRPO             & 56.54 & 90.43 & 77.07 & 80.06 & 45.02 & 73.55 & 16.67 & 14.00 & 55.54  \\
    \bottomrule
  \end{tabular}}
\end{table*}

\begin{figure}[t]
  \centering
\includegraphics[width=0.45\textwidth]{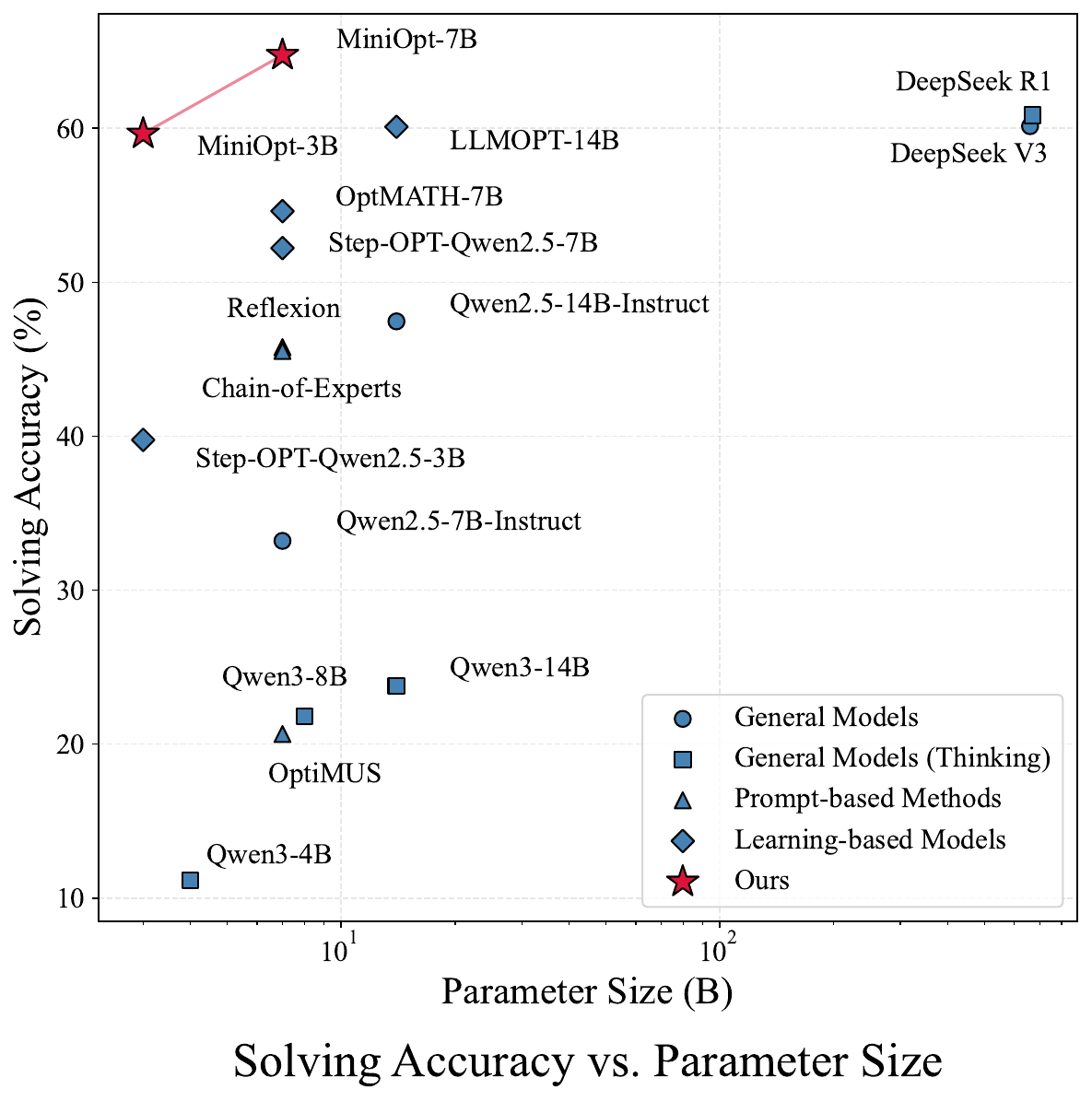}
  \caption{Comparison of average SA against model parameter scales for various methods. MiniOpt is the Pareto optimal among methods (with open-weight models).}
\label{fig:model_params_vs_performance_sa}
\end{figure}

\subsection{Pareto Front of Performance vs. Cost}

\textbf{Analysis of the Pareto Front (Answer to Q2).}
Figure~\ref{fig:model_params_vs_performance_sa} indicates that the MiniOpt family (represented by the solid red line) lies on the empirical Pareto frontier among the compared open-weight methods in the performance-versus-cost trade-off. Since the parameter size of GPT-5 and Gemini-2.5-Pro have not been disclosed, we do not label these two models in the figures.

As the scale of the model increases, the average SA performance of MiniOpt also grows steadily. It achieves a comprehensive performance lead while having substantially fewer parameters than top-tier general reasoning models such as DeepSeek-R1 and learning-based model LLMOPT-14B. Compared to the similar modeling and solving model Step-OPT, Step-OPT-Qwen2.5-7B achieves an average SA of 52.22\% across 8 benchmarks, while the proposed MiniOpt attains an average SA of 64.76\%. When the parameter scale of both models is reduced to 3B, Step-OPT-Qwen2.5-3B exhibits a performance drop of 12.46\%, while MiniOpt only decreases by 5.11\%. This indicates that the key advantage of MiniOpt lies in its ability to maintain superior performance even with a reduced parameter scale. From the perspective of capability density \cite{Xiao2025densing}, MiniOpt effectively achieves lower parameter requirements and inference costs while preserving comparable performance.

\subsection{Ablation Study}
\label{ablation_section}
We ablate core components of MiniOpt-3B and report results of the SA metric in Table~\ref{tab:ablation_sa} and the ER metric in Table S4 in supplementary material, where ``w'' denotes ``with'' and ``w/o'' denotes ``without''. As evidenced in Table~\ref{tab:ablation_sa}, each module of the proposed reasoning-to-model-and-solve paradigm demonstrates substantial contributions to model and solve optimization problems with smaller-scale models under limited training resources. Among these, 

\textbf{Importance of the Training Pipeline of MiniOpt (Answer to Q3).}
In the training pipeline of MiniOpt, each module plays a distinct role in improving SA and ER. First, the SFT warm-up of mid-training phase provides a better starting point for RL training. Without it, averages for SA fall to 56.60\%. The decreases are $\Delta$SA=–3.05. 
Second, collapsing the two-stage RL removes the progressive training that first consolidates the paradigm (stage-1) and then targets generalization (stage-2), the average SA drop to 55.07\% and the average ER of 82.73\% ($\Delta$ SA=–4.58\%, $\Delta$ ER=–5.19\%). 
Finally, reverting our OptGRPO to the original GRPO further decreases both metrics, the average SA of 56.54\% and ER of 82.35\%. Together these changes of OptGRPO improve sample efficiency and training stability, which is critical for eliciting strong optimization generalization at small parameter scales.

\textbf{Importance of the OptReward (Answer to Q4).} RL provides the most significant improvement, highlighting the importance of the proposed paradigm. The reasoning-to-model-and-solve paradigm and OptReward together turn free-form generation into a verifiable formulation that is easy to verify. Removing the OptReward and keeping only a final-answer signal (w/o OptReward) drops the average to 57.96\% and 85.31\% ($\Delta$ SA=–1.69\%, $\Delta$ ER= –2.61\%), respectively. The largest losses appear where the problems are challenging: Mamo Complex ($\Delta$ SA = –8.53\%, $\Delta$ ER = -8.52\%). These patterns align with the roles of the three reward components: the format score enforces the response contains complete and properly structured \texttt{<think>} and \texttt{<answer>} sections; the five-element score shapes the intermediate blueprint, so the model learns to extract problem structure before coding; the Accuracy Score certifies numerical correctness by executing the Pyomo code and comparing the returned optimum with the reference. Although the absolute performance gain of OptReward is smaller than that of the training pipeline, OptReward is indispensable because it provides the verifiable supervision signal that enables RL training without chain-of-thought annotations.

\begin{table*}[t]
  \centering
  \scriptsize
  \setlength{\tabcolsep}{3pt}
  \caption{Comparison of the SA metric between MiniOpt-1.5B and larger scale counterparts across 8 benchmarks.}
  \label{tab:small_sa}
  \resizebox{0.99\linewidth}{!}{
  \begin{tabular}{l l c| c c c c c c c c }
    \toprule
    \multicolumn{2}{l}{Solving Accuracy (SA)}  & \textbf{Avg.} & \textbf{NL4Opt} & \textbf{ICML.C} & \textbf{Mamo.E} & \textbf{Mamo.C} & \textbf{NLP4LP} & \textbf{Com.OR} & \textbf{Indus.OR} & \textbf{OptiBench}  \\
    \midrule
    \multicolumn{2}{l}{Number of Samples in Datasets} &  & 230 & 410 & 652 & 211 & 242 & 18 & 100 & 605 \\
    \midrule
    \multirow{3}[2]{*}{Ours}
        &MiniOpt-1.5B   & {50.33}   &88.7&	72.2&	78.99	&8.06&	71.07	&16.67&	16	&50.91 \\
        & MiniOpt-3B  & {59.65}   & {93.04} & {78.05} & 84.36 & {47.39} & {74.38} &  27.78& 17.00 & 55.21 \\
        &MiniOpt-7B   & {64.76}    &95.22  &   83.41   &89.57&51.18      &74.79&38.89&25.00&60.00\\
    \bottomrule
  \end{tabular}
  }
\end{table*}

\subsection{Mechanistic Analysis of Optimization Generalization (Answer to Q5)}
\label{sec:param_decoupled}

The previous experiments evaluate every method under standard optimization benchmarks. These benchmarks provide an end-to-end evaluation of optimization systems, where problem understanding, parameter extraction, optimization modeling and solving are jointly evaluated. While such an end-to-end evaluation reflects practical deployment scenarios, it does not reveal which capability primarily contributes to the observed performance improvements. In most existing optimization benchmarks, numerical parameters are directly embedded within the natural language problem description. To better understand where the learning-based methods obtains those optimization generalization ability and where its bottlenecks lie, we further introduce a complementary parameter-decoupled evaluation. 

\textbf{Experimental Setup and Inference Protocol.}
We construct OptiBench-Separate, a variant of the OptiBench benchmark~\cite{yang2025optibench} where tabular numerical parameters are externalized as CSV files. The data processing process is detailed in the supplementary material.

To ensure a fair comparison, we reconstruct the inference pipeline for all evaluated methods using the same two-stage inference protocol. Specifically, the model first analyzes the problem description to identify whether external parameter files are required and extracts the corresponding file names together with the associated schema information (e.g., column names). The extracted metadata is then incorporated into the subsequent code generation stage, where the model generates optimization code that explicitly loads the required data via \texttt{pd.read\_csv} and instantiates the optimization model through structured data access operations (e.g., \texttt{df.loc} and \texttt{df.iloc}) instead of directly relying on numerical values embedded in the prompt. The prompt of this pipeline is provided in supplementary material.

\textbf{Understanding Optimization Generalization through Parameter-Decoupled Evaluation.}
We evaluate five methods under this protocol: the proposed MiniOpt-3B, the general-purpose DeepSeek-V3 (671B, accessed via API), the learning-based models LLMOPT-Qwen2.5-14B, StepOPT-Qwen2.5-73B, and StepOPT-Qwen2.5-7B. 


\begin{table*}[t]
\caption{Parameter-decoupled evaluation on OptiBench-Separate (604 problems). All models evaluated under the same input format and scoring protocol.}
\label{tab:param_decoupled}
\centering
\small
\begin{tabular}{lccc}
\toprule
\textbf{Model} & \makecell{SA on \\ OptiBench} & \makecell{SA on \\ OptiBench-Separate} & $\Delta$SA \\
\midrule
DeepSeek-V3 (671B) & 64.13\% & 49.09\% & \textbf{-15.04\%} \\
LLMOPT-Qwen2.5-14B & 55.83\% & 25.62\% & -30.21\% \\
StepOPT-Qwen2.5-7B & 48.76\% & 28.93\% & -19.83\%\\
StepOPT-Qwen2.5-3B & 40.00\% & 16.36\% & -23.64\%\\
MiniOpt-3B & 55.21\% & 38.35\% & \textbf{-16.86\%}\\
\bottomrule
\end{tabular}
\end{table*}


Table~\ref{tab:param_decoupled} reports the performance of different methods under the proposed parameter-decoupled evaluation protocol. 
First, DeepSeek-V3 shows an absolute performance drop of 15.04\%, indicating that the absence of parameters substantially increases the difficulty of optimization modeling, even for large-scale general-purpose LLMs. Therefore, the organization of optimization data itself has a non-negligible impact on optimization reasoning, suggesting that optimization generalization depends not only on mathematical modeling, but also on how optimization instances are represented. Second, despite containing only 3B parameters, MiniOpt exhibits a performance degradation (16.86\%) that is comparable to that of the 671B DeepSeek-V3 (15.04\%), while consistently achieving a substantially higher solving accuracy than other learning-based optimization models. \textbf{This observation suggests that the proposed reasoning-to-model-and-solve paradigm combined with the RLVR framework effectively encourages structure-oriented optimization learning}. Third, among all learning-based optimization models, MiniOpt achieves both the smallest performance degradation and the highest solving accuracy on OptiBench-Separate. Compared with previous optimization-specialized models trained primarily through supervised fine-tuning, the proposed reasoning-to-model-and-solve paradigm explicitly reinforces both the completeness of optimization formulations and correctness of the solution. Consequently, MiniOpt encourages structure-oriented optimization learning, making the policy less dependent on parameter-coupled representations and leading to superior optimization generalization under parameter-decoupled evaluation.


\textbf{Error Pattern Analysis of Learning-based Optimization Models.}
To better understand why parameter decoupling degrades the performance of learning-based optimization models, we further conduct an error pattern analysis using MiniOpt as a representative case study.

Among the 605 optimization problems, 159 instances are correctly solved under the original OptiBench benchmark but become incorrect after parameter decoupling, accounting for the majority of the observed performance degradation. These instances therefore provide a direct window into how separating optimization structures from numerical parameters influences optimization modeling.
A closer inspection reveals that the dominant error pattern is not failure to recognize the existence of external parameter files. Instead, \textbf{the optimization models often correctly attempt to access external data, yet still formulate optimization models using hallucinated numerical coefficients rather than consistently grounding the optimization formulation on the retrieved parameters}. The dominant failure mode (57.9\%) correctly generates data-loading operations such as \texttt{pd.read\_csv}, this phenomenon suggests that existing learning-based optimization models are implicitly influenced by parameter-coupled training data. Under such a training paradigm, numerical parameters become a confounding factor during optimization learning.
Since optimization structures and parameter values always co-occur, models may partially rely on memorizing or reproducing numerical patterns while learning optimization formulations, making it difficult to distinguish optimization structure understanding from parameter processing. Consequently, once the numerical parameters are separated from the problem description, this hidden dependency is exposed, leading to degraded optimization performance.

From another perspective, these observations also highlight an overlooked data engineering issue in current optimization benchmarks. Existing optimization datasets typically couple optimization structures with instance-specific numerical parameters within a single textual prompt, rather than employing a pipeline. Although this representation is convenient for end-to-end evaluation, it differs from practical optimization systems, where optimization formulations and numerical data are naturally separated, with instance parameters commonly stored in structured files such as CSV or databases. Consequently, \textbf{parameter-coupled benchmarks inevitably entangle optimization structure understanding with parameter extraction}, making it difficult to independently assess optimization generalization.
Our case study suggests that the observed performance degradation is not primarily caused by the inability to process external parameter files, but by a hidden confounding effect introduced by parameter-coupled optimization datasets, where optimization structures and numerical parameters are inseparable during learning.

\section{Discussions}
\label{discussion}

\begin{figure}[t]
  \centering
\includegraphics[width=0.49\textwidth]{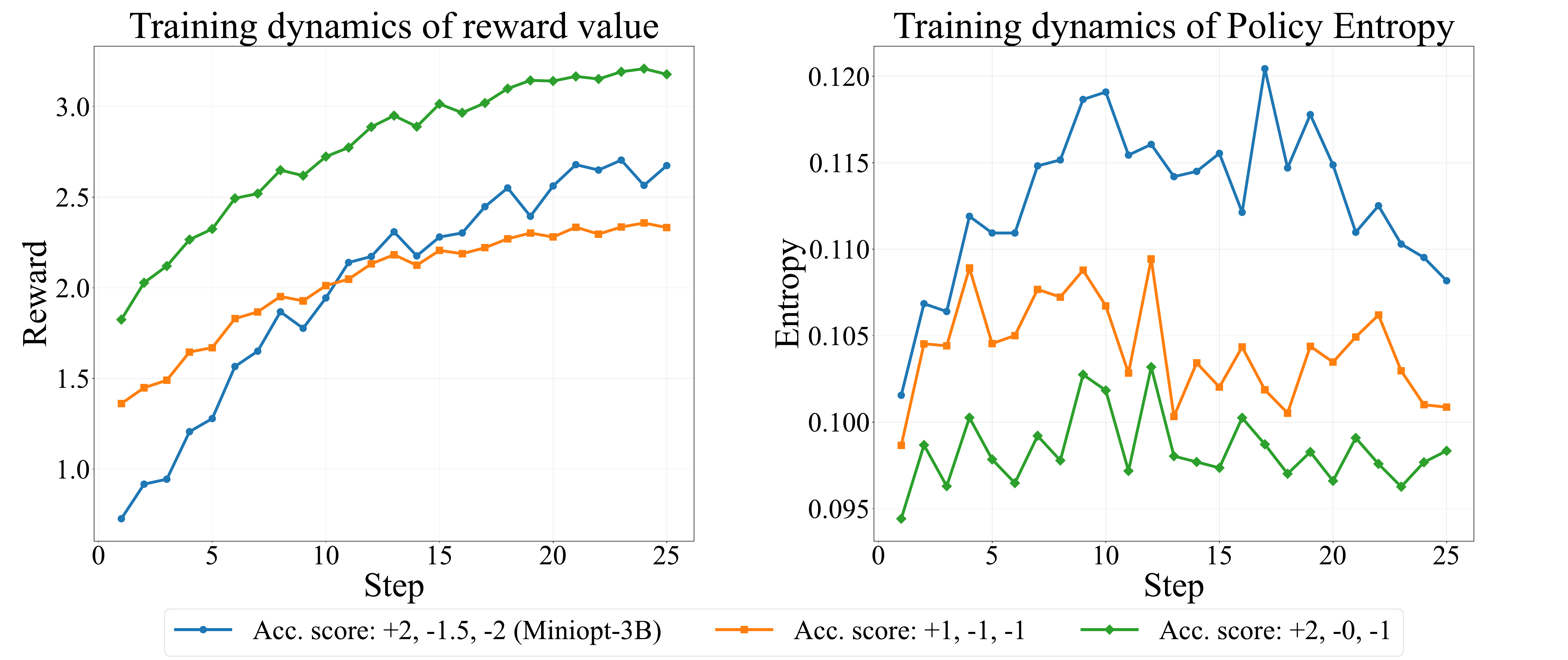}
    \caption{Comparison of training dynamics of reward values and policy entropy for MiniOpt-3B under different reward function settings during the second stage of RL training.}
    \label{fig:dynamics}
\end{figure}

\textbf{Sensitivity Analysis of the Accuracy Reward Design.} To investigate whether the effectiveness of OptReward depends on a specific reward assignment, we replace the original accuracy reward setting ((2,-1.5,-2)) with two alternative configurations, namely ((2,0,-1)) and ((1,-1,-1)). Figure~\ref{fig:dynamics} illustrates the training dynamics in terms of reward convergence and policy entropy, while Table~\ref{tab:ablation_score} reports the final solving accuracy (SA).
First, the ((2,0,-1)) setting achieves a noticeably higher converged reward than the original design, while yielding nearly identical SA. The increase in converged reward should not be interpreted as an improvement in optimization capability. Since incorrect solutions receive a neutral reward instead of a penalty, the expected reward distribution is shifted upward. Consequently, higher reward values are partially caused by reward rescaling rather than improved policy quality. The most significant performance drop is observed under the ((1,-1,-1)) setting, where SA decreases by 2.25\%. These results demonstrate that the effectiveness of OptReward relies on maintaining a hierarchical reward structure that progressively differentiates execution failures, incorrect solutions, and correct solutions. Such a design provides informative intermediate supervision and enables stable policy optimization, which is crucial for achieving strong optimization generalization under limited training resources. For more detailed analysis, please refer to the supplementary materials.

\begin{table*}[t]
\caption{Sensitivity analysis of the accuracy reward design. $\Delta_1$ denotes the reward gap between correct and incorrect solutions, while $\Delta_2$ denotes the reward gap between incorrect solutions and execution failures.}
\label{tab:ablation_score}
\centering
\small
\begin{tabular}{lccccc}
\toprule
\textbf{Reward Setting} & $\Delta_1$ & $\Delta_2$ & \textbf{Reward} & \textbf{Entropy} & \textbf{SA (\%)} \\
& $(R_c-R_i)$ & $(R_i-R_f)$ & \textbf{Convergence} & \textbf{Convergence} & \\
\midrule
Original $(2,-1.5,-2)$ & 3.5 & 0.5 & 2.65 & 0.110 & 59.65 \\
Variant A $(2,0,-1)$ & 2.0 & 1.0 & 3.20 & 0.097 & 59.36 \\
Variant B $(1,-1,-1)$ & 2.0 & 0.0 & 2.35 & 0.101 & 57.40 \\
\bottomrule
\end{tabular}
\end{table*}

\begin{figure}[t]
  \centering
\includegraphics[width=0.49\textwidth]{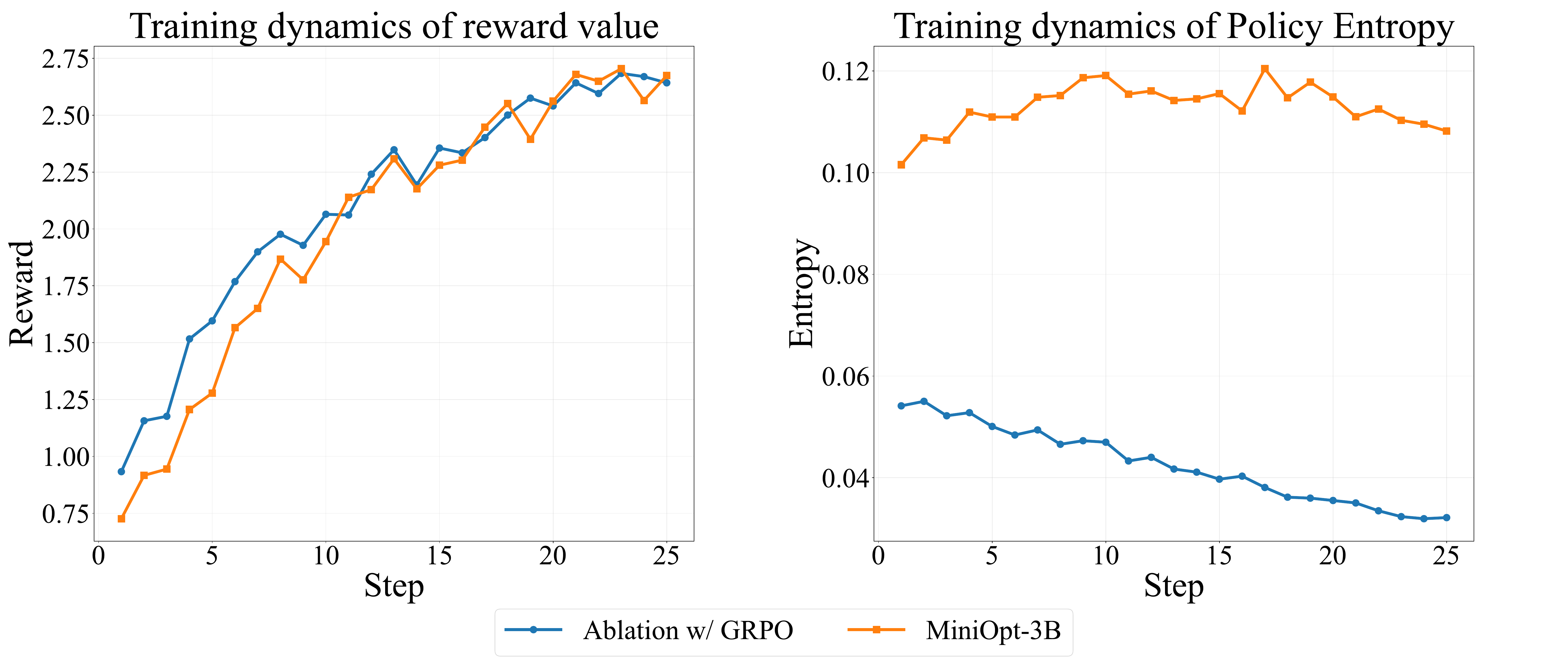}
    \caption{Comparison of training dynamics of reward function values and policy entropy between GRPO and OptGRPO during the second stage of RL training.}
    \label{fig:reward_entropy}
\end{figure}

\textbf{Ablation on OptGRPO and Training Dynamics Analysis.}
Empirical results (Section \ref{ablation_section}) demonstrate that OptGRPO consistently yields higher SA on downstream benchmarks compared to the vanilla GRPO. To better understand the reason for this improvement, we monitor the training dynamics by tracking both the reward function values and the policy entropy throughout the RL training process, as illustrated in Figure~\ref{fig:reward_entropy}.

While both GRPO and OptGRPO eventually converge to comparable reward values, their entropy trajectories diverge significantly. Standard GRPO exhibits a precipitous drop in entropy, indicative of premature policy collapse. In contrast, OptGRPO maintains a higher policy entropy throughout training, which encourages exploration of diverse reasoning trajectories. Optimization problems often possess complex, non-convex search spaces where multiple valid modeling paradigms exist. A premature drop in entropy restricts the policy to a narrow set of high-probability, yet potentially sub-optimal, reasoning trajectories. We hypothesize that maintaining higher policy entropy contributes to stronger optimization generalization by encouraging exploration of diverse reasoning trajectories.

\textbf{Exploration of Models with Smaller (Fewer than 3B) Parameter Scale.}
To further explore the performance of MiniOpt with smaller (fewer than 3B) parameter scales, thereby better balancing the trade-off between parameters and performance, we deploy MiniOpt-1.5B with the same {reasoning-to-model-and-solve} paradigm (Section~\ref{sec:method_paradigm}), OptReward (Section~\ref{sec:method_opt_reward}), and training pipeline (Section~\ref{sec:method_training_pipeline}). The results shown in Table \ref{tab:small_sa} and S5 in supplementary material show that when the model scale is reduced from 3B to 1.5B, MiniOpt achieves an average SA of 50.33\% and an average ER of 86.44\% across the 8 benchmarks, remains competitive with learning-based methods and general-purpose models with substantially larger parameter scales.

\textbf{Dimensionality Scalability Issue.}
Our experimental results show that \textit{MiniOpt-3B can successfully formulate optimization problems whose parameter dimensionality reaches 80. Although such dimensionalities are still modest from the perspective of large-scale optimization, they already pose substantial reasoning and modeling challenges for compact language models}. This shows a remarkably strong performance for models of such limited scale, which highlights that MiniOpt paradigm effectively enhances the model's optimization modeling capacity and retains robust generalization even in high-dimensional settings. For problems with even higher dimensionality, the descriptions of numerical parameters typically become substantially longer. Such extended input sequences inherently pose a significant challenge to the comprehension capabilities of small-scale models.

\textbf{Structural Scalability Beyond Context Length.}
The dimensionality analysis above reveals an important limitation of conventional parameter-coupled evaluation. When numerical parameters are directly embedded in natural-language descriptions, increasing the problem dimensionality inevitably increases the input length, making larger instances increasingly difficult to accommodate within the natural-language input.
To investigate this question, we evaluate MiniOpt under the parameter-decoupled setting of MIPLIB-NL~\cite{li2026constructing}, an industrial-scale optimization modeling benchmark reverse-constructed from MIPLIB 2017. MIPLIB-NL adopts a problem--data separation format, making it particularly suitable for evaluating structural scalability. We modify the inference procedure accordingly. The corresponding prompt for this process can be found in the supplementary materials. MiniOpt first identifies the external data files and their schemas, then reconstructs the optimization model from the problem description, and finally generates executable solver code that loads the numerical data from the corresponding files. Under this protocol, MiniOpt-3B successfully solves the instance which contains 1,620 binary decision variables and 3,285 constraints. MiniOpt-7B successfully solves instances with more than 1,620 variables and 2,520 constraints. Notably, the largest successfully solved instance is not monotonic with model scale, suggesting that solvability is determined not only by model size but also by the structural complexity and constraint interactions of individual instances.

\section{Conclusion}
\label{conclusion}
This paper presents MiniOpt, an optimization-specialized LLM that improves optimization generalization through a reasoning-to-model-and-solve paradigm and reinforcement learning from verifiable rewards. Unlike conventional supervised optimization modeling methods, MiniOpt encourages structure-oriented optimization learning by reinforcing the completeness of the optimization formulation and the correct solution. Extensive experiments on eight optimization benchmarks demonstrate that MiniOpt achieves competitive or superior optimization performance with substantially fewer parameters, lower training cost, and reduced inference overhead. Beyond end-to-end evaluation, we introduce a complementary parameter-decoupled evaluation protocol that disentangles the understanding of the optimization structure from numerical parameter processing. The resulting analyzes reveal that current learning-based optimization models are still affected by parameter-coupled data representations, while MiniOpt exhibits substantially stronger optimization generalization under parameter-decoupled settings, suggesting that small optimization-specialized LLMs possess the capacity to outperform conventional small optimization models and that their apparent scalability limits depend strongly on how optimization instances are represented.

\section*{Supplementary Material}
In the supplementary material, we provide all prompts used during the training of MiniOpt and all experiments, along with details on the construction of the training set and OptiBench-Separate. Furthermore, we include information on the benchmarks and baselines involved in the experiments, as well as additional results covering executability metrics, hyperparameter sensitivity analysis, experiments with varying MiniOpt parameter sizes, inference costs, and evaluations of general capabilities of MiniOpt.

\bibliographystyle{IEEEtran}
\bibliography{IEEEexample}

\includepdf[pages=-]{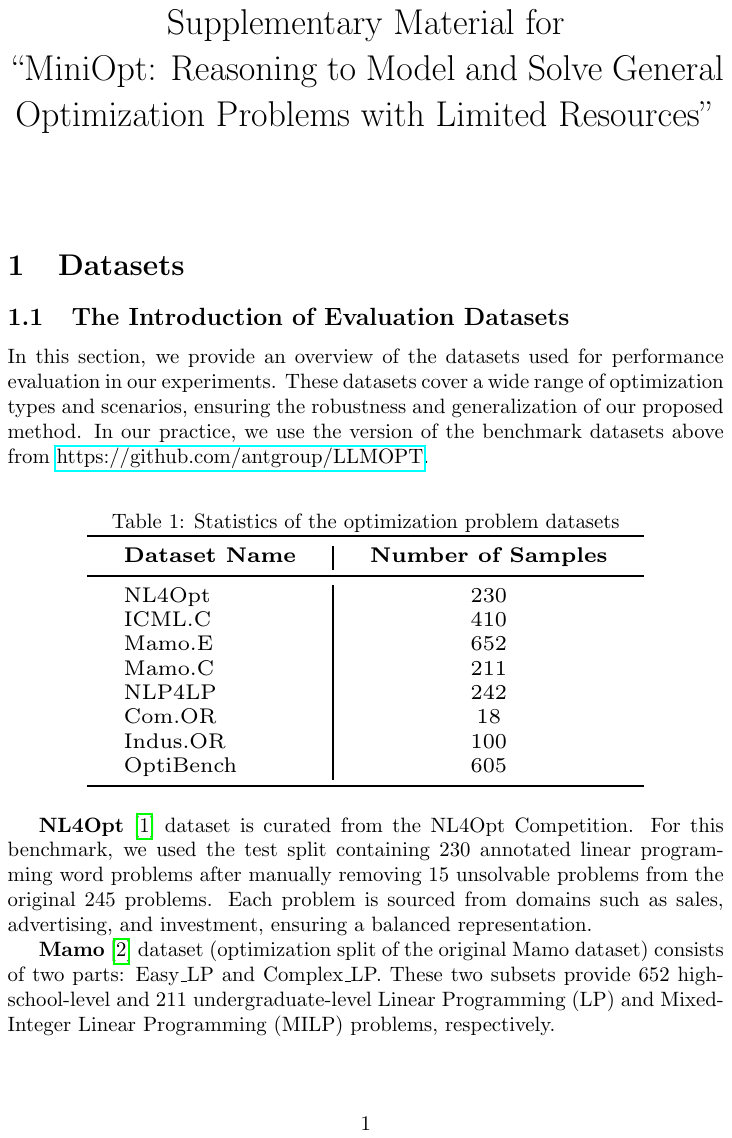}

\end{document}